\documentclass[conference]{IEEEtran}
\IEEEoverridecommandlockouts

\usepackage{amsmath}
\usepackage{amssymb}
\usepackage{cite}
\usepackage{url}
\usepackage{booktabs}
\usepackage{tabularx}
\usepackage{algorithm}
\usepackage{algpseudocode}
\usepackage{xcolor}
\usepackage{tikz}
\usetikzlibrary{positioning,arrows.meta,fit,calc}

\usepackage{enumitem}

\newcolumntype{Y}{>{\raggedright\arraybackslash}X}

\begin{document}

\title{Beyond Scene Description: Multi-Agent Orchestration for Non-Visual Access to Virtual Worlds}

\author{
	\IEEEauthorblockN{
		Toqeer Ali Syed\textsuperscript{1},
		Ali Akarma\textsuperscript{1,2}*,
		Adeel Ahmad\textsuperscript{1} and
		Danial Hameed\textsuperscript{3}
	}
	\IEEEauthorblockA{
		\textsuperscript{1}AI Center, Faculty of Computer and Information Systems,
		Islamic University of Madinah, Madinah, Saudi Arabia\\
		\texttt{toqeer@iu.edu.sa} \quad
		\texttt{443059463@stu.iu.edu.sa} \quad \\
		\texttt{443057803@stu.iu.edu.sa}\\
		\textsuperscript{2}AI V\&V Lab, King Fahd University of Petroleum and Minerals,
		Dhahran, Saudi Arabia\\
		\textsuperscript{3}Faculty of Computing and Information Technology,	University of the Punjab, Lahore, Pakistan\\
		\texttt{bsdsf23a014@pucit.edu.pk}\\
		*Corresponding author: \texttt{443059463@stu.iu.edu.sa}
	}
}

\maketitle

\begin{abstract}
Virtual worlds now host classrooms, meetings, conferences, shops, and social venues, and nearly every interaction they expose assumes a user who can scan a three-dimensional scene, follow avatars, and read floating panels. Blind and visually impaired (BVI) users are left with assistive tools that each solve one task in isolation: naming an object, reading text, describing a scene, or planning a route. A live virtual room defeats that model: obstacles, speakers, gestures, chat, slides, and notifications arrive together, and a tool that narrates all of them trades a visual barrier for an auditory one. This paper presents MetaBlind, an architecture that distributes nonvisual access across eight specialized agents, spanning perception, navigation, social and object interaction, communication, safety and trust, memory, and personalization, and that places an Accessibility Orchestrator between those agents and the user. Agents publish candidate information into a shared accessibility context instead of speaking to the user directly. The orchestrator scores each candidate on safety relevance, goal relevance, urgency, confidence, user relevance, and estimated listening load, then releases only the items it judges relevant at that moment through speech, structured audio, or haptic output. We give the selection step a formal statement, specify the orchestration cycle as an algorithm, and define an evaluation protocol against a single-agent assistant. MetaBlind is reported at the design stage, with no prototype measurement or user study, and the protocol states which outcomes would support the design and which would refute it.
\end{abstract}


\section{Introduction}

Immersive platforms are no longer limited to entertainment. Universities run lectures in virtual halls, employers hold team meetings in virtual offices, and clinicians rehearse procedures in simulated wards. Access to these platforms therefore bears on participation in education and work, where agentic and federated systems are increasingly recognized as vital for disability-inclusive employment and collaborative equity~\cite{syed2026fedagent}.

The interaction model behind them stays visual. A user is expected to read a room layout at a glance, tell avatars apart, notice a raised hand, and walk a path around furniture. Each of those actions maps onto vision and onto the spatial reasoning that vision supports.

Blind and visually impaired (BVI) users meet a different environment. Screen readers handle the parts of an interface that are declared as text and structure, and they say little about a room whose geometry, occupancy, and activity change from second to second. Camera-based assistants describe what sits in front of the user at one instant, which suits a photograph or a product label. Neither one addresses the property that makes a virtual room difficult to convey nonvisually: several channels of information stay active at once, and their relative importance shifts with what the user is doing.

Large language models (LLMs), vision-language models (VLMs), and agentic systems that plan, call tools, and hold state change what an accessibility layer can attempt. The obvious application is a single strong assistant that describes everything it sees. We argue against that design. A monolithic describer reproduces the problem that verbose audio guidance already creates: speech fills the channel the user also needs for the lecturer, for a colleague, and for spatial audio cues. The scarce resource in an immersive setting is not detection but the user's attention.

MetaBlind treats nonvisual access as a coordination problem. Responsibility is split across agents with narrow scopes, each of which can hold its own state and call its own tools. None of them talks to the user. They write candidate information into a shared accessibility context, and an Accessibility Orchestrator decides what reaches the user, in what order, and through which modality. Safety-critical guidance preempts description. Description that does not serve the current task waits or is dropped.

The status of the work is plain. This paper reports no measurements, every claim about behavior is a design argument rather than an experimental result, and a longer treatment of the same project is in preparation.

\subsection*{Contributions}

\begin{enumerate}
\item An orchestrated multi-agent architecture for BVI participation in immersive virtual environments, with a shared accessibility context as the only channel between specialized agents and the output layer.
\item A statement of the orchestration decision as a scored, load-constrained selection over candidate items, with safety relevance and estimated listening load as explicit terms.
\item An orchestration cycle that specifies where safety gating, conflict handling, scheduling, and memory update sit relative to agent execution.
\item An evaluation protocol that compares the architecture against a non-orchestrated single-agent assistant on task outcomes, communication economy, and subjective measures, together with the outcomes that would count as evidence against the design.
\end{enumerate}

\section{Problem Formulation}

\subsection{Setting and Assumptions}

Consider one BVI user inside a shared virtual environment. At time $t$ the environment has a state $E_t$ that covers geometry, static and interactive objects, avatars and their activity, text and interface surfaces, chat traffic, and system events. The architecture assumes read access to $E_t$, whether through platform instrumentation, scene graph queries, rendered views passed to a VLM, or a combination. That assumption constrains deployment, and we return to it in Section~\ref{sec:limits}.

The user carries a profile $U$ holding preferences over description detail, navigation style, speech verbosity, object categories of interest, and output modality. Clinically, BVI individuals exhibit substantial heterogeneity in residual visual acuity, refractive status, and functional reading distance~\cite{Yam2023,NonCyclo2025}, as cataloged across standardized vision screening guidelines~\cite{PreventBlindness} and validated clinical acuity assessments~\cite{Roberts2026,Du2026}. Capturing these parameters within $U$ ensures that the assistive system can accommodate users who retain partial functional sight alongside those who are totally blind. The user pursues a goal $G_t$, which may be explicit, as in a spoken request, or inferred from recent activity.

\subsection{Shared Accessibility Context}

All agents read and write one structure, the shared accessibility context $C_t$. It holds the current interpretation of the environment: recognized objects and their positions, identified avatars and speaking state, the active route, the last items delivered to the user, and open user requests. The context exists so that agents reason about the same room rather than about private and possibly inconsistent snapshots. Conceptually, this shared state functions as an agentic digital twin of the virtual room, maintaining a synchronized, auditable representation of dynamic surroundings to ground multi-agent coordination~\cite{syed2026climate,syed2026agenticdt}. A navigation decision and a social description then rest on the same belief about where the user stands and who is nearby.

\subsection{Candidate Information and Selection}

The specialized agents form the set
\begin{equation}
\mathcal{A}=\{A_p,A_n,A_s,A_o,A_c,A_t,A_m,A_u\},
\end{equation}
for perception, navigation, social interaction, object interaction, communication, safety and trust, memory, and personalization. Each agent emits a candidate item
\begin{equation}
I_i = A_i(E_t,U,C_t),
\end{equation}
where a candidate may be an obstacle warning, a turn instruction, a speaker attribution, an object affordance, an incoming message, or a structured description of a diagram.

The orchestrator does not forward the full candidate pool $\mathcal{I}=\{I_1,\dots,I_n\}$. It selects a subset
\begin{equation}
I^{*}=\operatorname{Select}\bigl(I_1,\dots,I_n \mid G_t,U,C_t,R\bigr),
\end{equation}
with $R$ carrying risk and urgency. Each candidate receives a priority
\begin{equation}
P(I_i)=w_1R_i+w_2G_i+w_3T_i+w_4C_i+w_5U_i-w_6L_i,
\label{eq:priority}
\end{equation}
where $R_i$ is safety relevance, $G_i$ relevance to the current goal, $T_i$ temporal urgency, $C_i$ the confidence reported by the producing agent, $U_i$ relevance under the user profile, and $L_i$ an estimate of the cognitive and auditory load the item imposes. The load term enters with a negative weight, so a long description has to earn its delivery.

\subsection{Delivery as a Budgeted Choice}

Equation~\eqref{eq:priority} ranks candidates but does not say how many of them to deliver. We state the decision as a choice under a load budget. Let $B(U,G_t)$ be the load the user tolerates in the current window, derived from the profile and the task, and let $\mathcal{S}$ be the set of items the safety gate marks as mandatory. The orchestrator solves
\begin{equation}
I^{*}=\operatorname*{arg\,max}_{X\subseteq\mathcal{I},\ \mathcal{S}\subseteq X}\ \sum_{I_i\in X}P(I_i)
\quad\text{s.t.}\quad \sum_{I_i\in X}L_i\le B(U,G_t).
\label{eq:budget}
\end{equation}
Three consequences follow. Mandatory safety items are never crowded out, because the constraint forces them into $X$. Items that score well on goal relevance but carry a heavy listening load now compete against cheaper items instead of being delivered by default. A quiet channel becomes a legitimate outcome, since an empty selection is feasible whenever no candidate earns its load.

The candidate pool in one cycle is small, so greedy selection by the ratio $P(I_i)/L_i$ is adequate at the rates an audio channel supports.

\subsection{Design Requirements}

The formulation above imposes seven requirements, and each one is assigned to a named element of the architecture in Section~\ref{sec:arch} rather than to the system as a whole.

\begin{enumerate}[label=R\arabic*.,leftmargin=*]
\item \textit{Coordination.} Heterogeneous agents act on one environment without contending for the output channel.
\item \textit{Shared representation.} Agents hold a common belief about the room, which $C_t$ provides.
\item \textit{Prioritization.} Candidates are ranked and filtered before delivery, through $P(I_i)$ and the budget in \eqref{eq:budget}.
\item \textit{Conflict resolution.} Contradictory candidates are detected and not delivered together.
\item \textit{Uncertainty.} Agent confidence reaches the delivery decision as $C_i$ rather than being discarded upstream.
\item \textit{Personalization.} How much is said, when, and through which channel follows the individual user through $U$ and $B$.
\item \textit{Privacy and safety.} High-impact guidance and sensitive requests pass a gate before they reach the user or the environment, adhering to multi-pillar ethical AI governance and trustworthy cybersecurity principles that require strict verification prior to autonomous action~\cite{jan2026eagf}.
\end{enumerate}

\section{Related Work}

\subsection{Task-Specific Assistance for BVI Users}

Assistive technology for BVI users has a long record in electronic travel aids, obstacle detection, optical character recognition, wearable cameras, indoor and outdoor guidance, and haptic displays. In parallel, advances in AI-driven ophthalmic screening, clinical triage, and automated reading assessment have demonstrated the capacity of machine learning to evaluate visual deficits and functional text processing~\cite{Somerville2026,Hariyama2026}. In assistive computing, two lines matter here. Visual question answering brought human and computational answers to photographs taken by BVI users \cite{bigham2010}, and the benchmark that followed exposed how far real BVI photographs sit from curated vision datasets \cite{gurari2018}. Work on navigation added collaborative structure, with sighted and blind participants building routes together \cite{balata2014}, and crowdsourced route mapping carried indoor guidance into buildings that lack instrumentation \cite{plikynas2022}.

These systems share a common structure: each answers a single class of question about a mostly static scene, on the user's request. An immersive room asks a different question, because nothing waits for the request.

\subsection{Nonvisual Access to Immersive Environments}

Research on accessibility in virtual and augmented reality (VR/AR) has shown that nonvisual representations can improve access: audio description, spatial audio, sonification, haptic feedback, accessible menu structures, and alternative locomotion each restore part of an immersive experience without sight. In parallel, collaborative augmented reality (AR) and shared virtual frameworks have established methods for multi-user interaction and spatial asset management while maintaining data and operational integrity~\cite{syed2022car}.

The limitation lies in where the intelligence sits. Most of this work treats accessibility as interface translation, mapping visual content onto another channel. Translation says nothing about selection. A virtual classroom holds a lecturer, a dozen avatars, furniture, a slide deck, a chat panel, gestures, and notifications, and a faithful translation of all of it is unusable. The missing component decides which fraction of the room is worth a sentence right now.

\subsection{Vision-Language Models for Visual Assistance}

VLMs describe scenes, read text in context, and answer follow-up questions at a quality earlier pipelines did not reach, and they are now being applied to guidance for blind and low-vision users. Their known failure modes carry directly into immersive settings. A hallucinated object becomes a hallucinated obstacle. Weak spatial grounding becomes a wrong turn. Latency that a photograph tolerates is intolerable once the room has already changed. Excessive verbal description occupies the audio channel the user needs for speech and for spatial cues.

\subsection{Agentic and Multi-Agent Systems}

Agentic systems plan, call tools, keep state, and check their own intermediate results, which lets a designer assign narrow responsibilities to separate components. Across complex cyber-physical settings, agentic AI has been coupled with digital twins for secure, autonomous, and auditable spatial management and environment monitoring~\cite{syed2026climate,syed2026agenticdt}. Furthermore, agentic architectures have demonstrated proficiency in solving multi-objective decision-making and planning tasks under strict resource and budget constraints~\cite{syed2025finagent,Syed2026FinNutriAgent}, showing how autonomous agents can balance competing user goals within bounded operational budgets.

Accessibility has begun to adopt the pattern. Beyond indoor navigation through agentic floor-plan parsing \cite{ayanzadeh2026}, secure federated and agentic frameworks have been introduced to advance disability-inclusive employment in intelligent environments~\cite{syed2026fedagent}. These systems establish that specialized agents can cooperate effectively on perception, reasoning, and assistive coordination, yet their application to real-time nonvisual arbitration in immersive 3D spaces remains an open frontier.

\subsection{The Gap}

Four bodies of work are relevant, and none of them meets the case at hand. Task-specific assistance answers one question at a time. Work on accessible VR/AR translates channels without deciding what to translate. VLMs interpret scenes and leave arbitration open. Agentic accessibility work has addressed navigation and has not been applied to sustained participation in a shared virtual world, where perception, movement, conversation, object manipulation, and messaging all make claims on the user at once.

The unresolved problem is arbitration: given several agents that each hold a correct and useful observation, what reaches the user, in what order, and what is dropped without comment. A navigation agent with an obstacle, a social agent with a speaker attribution, and a perception agent with a new slide are all right, and delivering all three together is wrong. MetaBlind is a proposal for that decision.

\section{The MetaBlind Architecture}
\label{sec:arch}

\subsection{Structure}

\begin{figure*}[t]
\centering
\includegraphics[width=\textwidth]{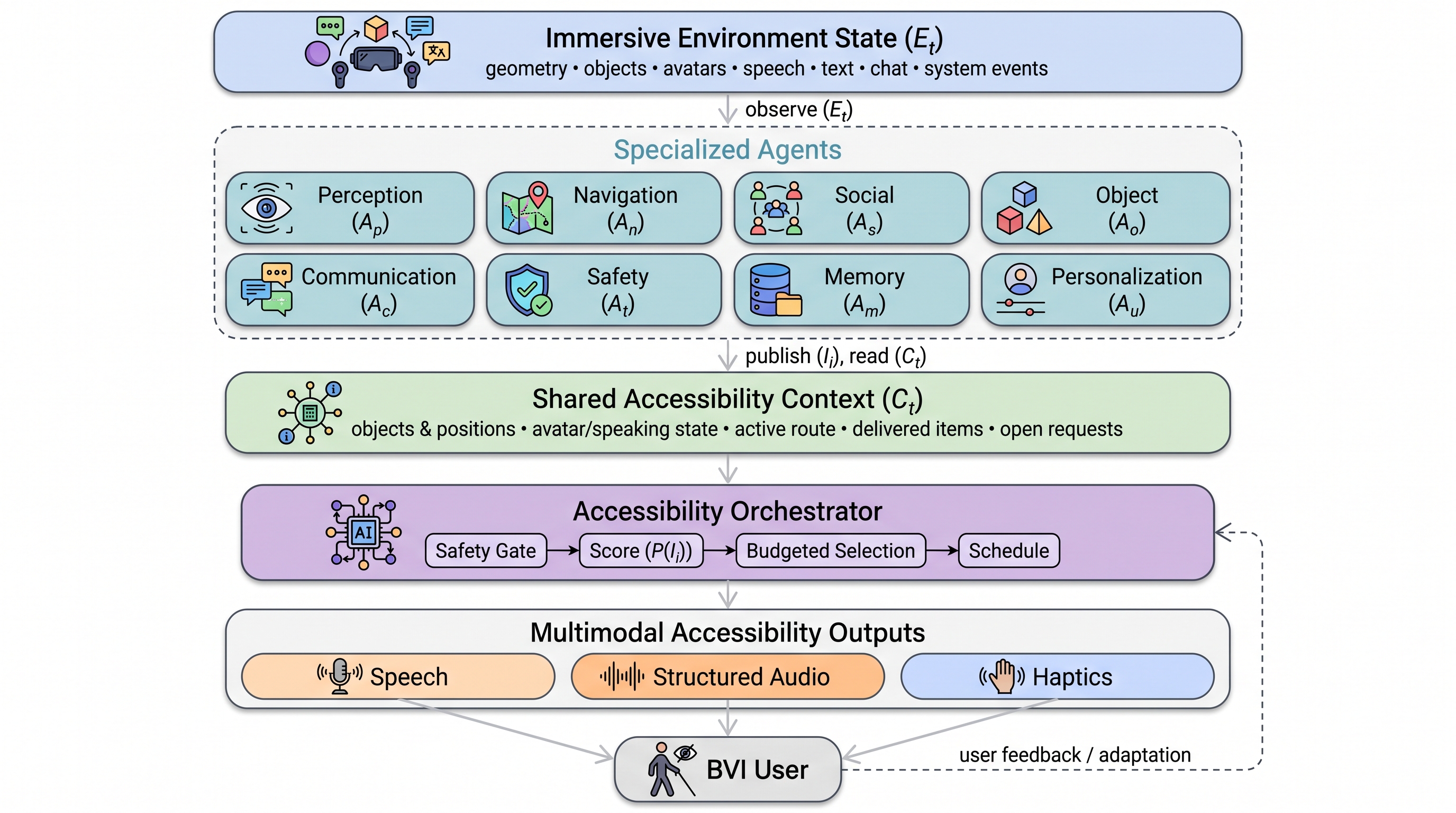}
\caption{The MetaBlind architecture. Specialized agents interpret the environment and publish candidate items into the shared accessibility context rather than addressing the user. The orchestrator holds the only path to the output layer, so it decides what the user hears and feels, and in what order. The return path on the right carries user queries, control actions, and feedback.}
\label{fig:arch}
\end{figure*}

Fig.~\ref{fig:arch} shows the arrangement. Agents observe the environment and write into the shared context, the orchestrator reads that context and drives the output layer, and user speech, queries, and control actions return through the orchestrator.

One design rule shapes everything else: no specialized agent owns the output channel. Direct agent-to-user speech is what produces overlap and contradiction, so the architecture removes the possibility.

Centralizing arbitration is a deliberate choice. Agents that negotiate among themselves would each need a model of the user's attention, and the cost of holding that state in eight places is duplicated logic and inconsistent behavior. One component that sees every candidate can compare them, which is what \eqref{eq:budget} requires. The price is a single point of failure and one more stage of delay, and Section~\ref{sec:limits} treats both.

\subsection{Specialized Agents}

Table~\ref{tab:agents} lists the agents, their scope, and the kind of candidate each contributes. Four points deserve emphasis beyond the table.

The Safety and Trust Agent is not one voice among many. It screens candidates before any high-impact instruction reaches the user, examining reported uncertainty, disagreement between agents, interactions that look manipulative, actions with consequences the user may not intend, and requests that touch private information. Its output is a gate, and it supplies the mandatory set $\mathcal{S}$ in \eqref{eq:budget}. Because agentic systems in immersive spaces ingest open multimodal streams and execute high-consequence actions, dedicated defenses are necessary to detect adversarial manipulation, malicious instructions, and prompt injection attacks before guidance is dispatched~\cite{syed2025toward}. In addition, safeguarding the underlying sensing pipeline against network-level disruptions, such as distributed denial-of-service (DDoS) traffic, is essential to preserve the availability and temporal integrity of real-time assistive streams~\cite{shaikh2024advancing}.

The Memory Agent prevents redundant description. A room the user has entered ten times needs no full account on the eleventh visit. In effect, memory lowers $L_i$ by shortening what has to be said.

The Personalization Agent supplies the shape of $U$ and the budget $B$. BVI users differ in what they want from an assistant, from terse landmark cues to detailed scene narration, and a single fixed verbosity setting serves neither end of that range.

Conflict handling is distributed rather than centralized. Confidence enters through $C_i$, which down-weights uncertain candidates without discarding them, and contradictions between agents are surfaced to the safety gate, which withholds high-impact guidance until the disagreement resolves. We claim no complete resolution policy here. Which policy performs best under time pressure is an open question, listed in Section~\ref{sec:limits}.

\begin{table}[t]
\caption{Agents, scope, and representative candidate output}
\label{tab:agents}
\footnotesize
\begin{tabularx}{\columnwidth}{@{}l Y@{}}
\toprule
\textbf{Agent} & \textbf{Scope and representative candidate} \\
\midrule
Perception $A_p$ & Objects, avatars, text surfaces, interface elements, spatial relations, environmental change. ``A whiteboard is on the wall to your left.'' \\
Navigation $A_n$ & Accessible paths, destinations, virtual obstacles, orientation. ``Two steps forward, then turn right.'' \\
Social $A_s$ & Nearby avatars, who holds the floor, turn-taking, relevant nonverbal events. ``Sara, on your right, is speaking to you.'' \\
Object $A_o$ & Which objects are interactive, what they do, mapping user intent onto a virtual action. ``The panel opens the shared notes.'' \\
Communication $A_c$ & Chat, announcements, meeting information, notifications, messages to other participants. ``A message from the host.'' \\
Safety and trust $A_t$ & Uncertainty, agent disagreement, suspicious interaction, unsafe action, privacy-sensitive request. Gate decision and mandatory set $\mathcal{S}$. \\
Memory $A_m$ & Visited locations, frequently handled objects, interaction history, approved preferences. Suppression of a known description. \\
Personalization $A_u$ & Detail level, navigation style, verbosity, object categories, modality. Parameters of $U$ and budget $B$. \\
\bottomrule
\end{tabularx}
\end{table}

\subsection{Orchestration Cycle}

Algorithm~\ref{alg:cycle} states one cycle. The order of operations carries the design commitments. Safety gating runs before scoring, so a mandatory item cannot be outscored. Selection runs against a budget, so silence stays reachable. Delivered items are written back into the context, so the next cycle knows what the user has already heard and does not repeat it.

The arbitration itself is inexpensive. One cycle invokes each agent once, evaluates $|\mathcal{I}|$ scores, and selects greedily in $O(|\mathcal{I}|\log|\mathcal{I}|)$, with $|\mathcal{I}|$ on the order of the number of agents. The model calls inside each $A_i$ dominate the cost of a cycle, which points to invoking agents selectively, driven by what changed in $E_t$ since the last cycle, rather than polling all eight on a fixed clock.

\begin{algorithm}[t]
\caption{One orchestration cycle at time $t$}
\label{alg:cycle}
\begin{algorithmic}[1]
\Require environment state $E_t$, context $C_t$, profile $U$, goal $G_t$
\State $\mathcal{I}\gets\emptyset$
\ForAll{$A_i\in\mathcal{A}$}
  \State $I_i\gets A_i(E_t,U,C_t)$ \Comment{candidates with confidence}
  \State $\mathcal{I}\gets\mathcal{I}\cup\{I_i\}$
\EndFor
\State $\mathcal{I}\gets\Call{Deduplicate}{\mathcal{I},C_t}$ \Comment{drop items just delivered}
\State $\mathcal{S},\mathcal{I}\gets\Call{SafetyGate}{\mathcal{I}}$ \Comment{mandatory set, gated pool}
\ForAll{$I_i\in\mathcal{I}$}
  \State $P(I_i)\gets w_1R_i+w_2G_i+w_3T_i+w_4C_i+w_5U_i-w_6L_i$
\EndFor
\State $B\gets\Call{Budget}{U,G_t}$
\State $I^{*}\gets\Call{Select}{\mathcal{I},\mathcal{S},P,L,B}$ \Comment{greedy on $P/L$, \eqref{eq:budget}}
\ForAll{$I_i\in I^{*}$ in order of $P$}
  \State \Call{Render}{$I_i$, modality from $U$}
\EndFor
\State $C_{t+1}\gets\Call{Update}{C_t,\mathcal{I},I^{*}}$
\end{algorithmic}
\end{algorithm}

\subsection{Output Layer}

Delivery spans speech, structured audio, and haptic devices when available. The choice of modality matters for the load term. An obstacle rendered as a short spatialized tone costs a fraction of the same warning rendered as a sentence, and it leaves the speech channel free for the lecturer. Modality assignment therefore belongs to the profile rather than to a global setting.

\subsection{Worked Example}

A blind student attends a virtual university lecture on cybersecurity. On entry, $A_p$ registers the room and the screen, $A_s$ reports who holds the floor and identifies two classmates nearby, and $A_n$ fixes the student's orientation. The orchestrator delivers an entry summary and then goes quiet.

A network topology diagram appears on the screen. Rather than reading the visible labels in raster order, $A_p$ produces a structured account of it: the segments, what connects them, and where the boundary devices sit. The student asks where the firewall is. That query raises $G_t$, and the orchestrator coordinates perception, object interaction, and navigation to answer in the diagram's own terms.

Mid-answer, a classmate raises a hand. $A_s$ has a valid candidate. It scores well on user relevance and poorly on goal relevance, and it is costly in the middle of a spatial explanation, so the orchestrator holds it and delivers it at the next gap. An obstacle on the student's path, by contrast, enters through the safety gate as mandatory and interrupts at once. The same architecture produces both behaviors, and the difference comes from the gate and the budget rather than from agent priority levels fixed in advance.

\section{Evaluation Protocol}
\label{sec:eval}

The architecture makes an empirical claim: coordinating specialized agents through a budgeted orchestrator serves BVI users better than a capable single assistant that describes what it detects. The protocol below is designed to test that claim. It has not yet been run.

\textbf{Environments.} We use four controlled immersive scenarios, chosen because they exercise different agents: a virtual classroom, a virtual conference, a virtual shopping environment, and a virtual workplace. Scripted events ensure that the same obstacle, the same overlapping speaker, and the same notification reach every participant at comparable points.

\textbf{Conditions.} MetaBlind is compared against a baseline single-agent accessibility assistant that shares the same underlying models, the same environment access, and the same output modalities, and that lacks the shared context, the safety gate, and the budgeted selection. Holding the model constant isolates orchestration from model quality. A further ablation keeps the agents and removes the budget, which separates the value of decomposition from the value of arbitration.

\textbf{Participants and tasks.} BVI participants perform per-scenario tasks with a verifiable end state: reach a named location, identify and operate a specific object, answer a question about a presented diagram, and respond to a participant who addresses the user by name.

\textbf{Measures.} The measures fall into three families. Task effectiveness covers the completion rate, the time to completion, navigation errors, accuracy in identifying objects, and successful social exchanges. Communication economy covers the number of accessibility messages a participant judges unnecessary, interruption frequency, and the delay between a triggering event and the matching delivery. Subjective measures cover cognitive and auditory workload, perceived independence, trust in the assistance, satisfaction, and overall usability.

\textbf{Outcomes that would refute the design.} Orchestration adds a decision layer, and the layer has costs. Three results would count against it. Delay from event to delivery may rise past what navigation tolerates, which would make the arbitration too slow for the setting it targets. The suppression policy may withhold information participants wanted, which would appear as task failures concentrated in the withheld categories. Participants may report equivalent workload across conditions, which would place any benefit in agent specialization rather than in orchestration. We regard the third outcome as the most plausible of the three, and the ablation is included to detect it.

\section{Discussion and Limitations}
\label{sec:limits}

No empirical validation appears in this paper. MetaBlind is a design and a formulation, and Section~\ref{sec:eval} states what would settle the question. Nothing here should be read as a performance result.

Environment access is the binding practical constraint. Equation~\eqref{eq:priority} assumes candidates grounded in $E_t$, and the quality of that grounding depends on what the platform exposes. A platform with a queryable scene graph supports reliable spatial reasoning. A platform that offers only rendered frames pushes the work onto a VLM and inherits its errors, which do more damage here than in a captioning system because the user acts on the output.

Deliberation competes with responsiveness. Every stage in Algorithm~\ref{alg:cycle} adds delay, and an obstacle warning that arrives late is not an accessibility feature. Selective agent invocation and cached context are the obvious mitigations, and their cost in accuracy is unmeasured.

Observation in shared spaces raises questions about other people. Agents that attribute speech, read gestures, and track who stands where are processing information about participants who did not consent to that processing and who may not know it is happening. Keeping the layer safe for the BVI user and acceptable to everyone else in the room is a requirement, and that dimension needs treatment of its own. Operationalizing these safeguards calls for comprehensive ethical AI governance frameworks~\cite{jan2026eagf} that enforce multi-stakeholder privacy boundaries, auditable decision logs~\cite{syed2026agenticdt}, and trustworthy security constraints before deploying autonomous assistants in populated social venues.

Four questions stay open. How should a shared context be represented so that heterogeneous agents stay consistent as a room changes? Which conflict-resolution policy performs best when agents disagree under time pressure? How should the weights in \eqref{eq:priority} and the budget $B$ be fit, per user, per task, or learned from interaction? Finally, does decomposition or arbitration carry the benefit, if a benefit exists?

\section{Conclusion}

Immersive platforms are becoming places where people study and work, and assistive tools built for single tasks do not carry over to them. MetaBlind proposes a different division of labor. Specialized agents interpret the environment and publish candidates into one shared context. An orchestrator holds the output channel and decides what the user hears, scoring candidates on safety, goal relevance, urgency, confidence, user preference, and listening load, then selecting under a load budget that makes silence a valid answer. The contribution is the arbitration rather than the detection.

The next step is a prototype and the study in Section~\ref{sec:eval}, with the ablation that separates agent specialization from orchestration, alongside a systematic review of accessible VR/AR, agentic accessibility, and VLM-based assistance for blind and low-vision users. Until those measurements exist, MetaBlind stands as an architecture with a stated formulation and a falsifiable claim.

\bibliographystyle{IEEEtran}
\bibliography{references}

\end{document}